\documentclass[conference]{IEEEtran}
\IEEEoverridecommandlockouts
\usepackage{cite}
\usepackage{amsmath,amssymb,amsfonts}
\usepackage{algorithmic}
\usepackage{graphicx}
\usepackage{textcomp}
\usepackage{xcolor}
\def\BibTeX{{\rm B\kern-.05em{\sc i\kern-.025em b}\kern-.08em
    T\kern-.1667em\lower.7ex\hbox{E}\kern-.125emX}}
\begin{document}

\title{Equilivest: A Robotic Vest to aid in Post-Stroke Dynamic Balance Rehabilitation
}

\author{%
Franco Paviotti$^{1}$,  Esteban Buniak $^{2}$, Rodrigo Ramele $^{3}$, Orestes Freixes$^{4}$ and Juan Miguel Santos$^{5}$
\thanks{$^{1}$F. Paviotti, $^{2}$E.Buniak and $^{3}$R.Ramele  are with the Instituto Tecnológico de Buenos Aires (ITBA)
                    Buenos Aires, Argentina
        {\tt\small \{fpaviotti|ebuniak|rramele\}@itba.edu.ar}}%
\thanks{$^{4}$O. Freixes is with the CINER Centro Integral de Neurorehabilitación,
                    Buenos Aires, Argentina
        {\tt\small orestesfreixes@gmail.com}}%
\thanks{$^{5}$J.M. Santos is with the UNAHUR Universidad Nacional de Hurlingham,
                    Buenos Aires, Argentina
        {\tt\small juan.santos@unahur.edu.ar}}%
\thanks{\textbf{This extended abstract was presented at the Workshop on Assistive Robotic Systems for Human Balancing and Walking: Emerging Trends and Perspectives at IROS2022.}}

}


\maketitle


\begin{IEEEkeywords}
Stroke, Dynamic Balance, Neurorehabilitation, Biofeedback, Vibrotactile
\end{IEEEkeywords}

\section{Introduction}

Brain stroke is a devastating medical condition, that affects world population and is the main cause of disabilities worldwide~\cite{Caplan.etal2023}.  Disabilities related to stroke can affect motor pathways, and may lead to several motor function disorders.   One important aspect of motor function is balance which is the ability to control the body's center of mass inside the base support provided by the lower limb~\cite{Bowman2021}.  Stroke can affect dynamic balance as well,  which is manifested while walking, impairing autonomy and independence, important factors in Activities of Daily Living (ADL) particularly for young patients~\cite{Afrasiabifar.etal2020,Donato.etal2016}.


Strong evidence suggests that neuroplasticity can be enhanced by neural rehabilitation~\cite{DeAngelis.etal2021,Albert.etal2012}.  These procedures are aimed to relearn movements that can trigger new neural pathway generation which reroute, or even completely replace, those pathways that were damaged by the stroke.   Recently, biofeedback techniques aiming to provide extra information to the patient to aid in the relearning, have appeared as a complementary treatment to increase neuralplasticity. These are in the form of Wearable devices-based biofeedback rehabilitation (WDBR)~\cite{Peake.etal2018} or robotic rehabilitation gait devices~\cite{Zhao.etal2022,Peshkin.etal2005,Tong.etal2006}.  



Therefore, the addition of an independent and new peripheral therapeutic signal, that can be assimilated as an extrasensory input~\cite{Brandebusemeyer.etal2021},  could improve dynamic balance performance on post-stroke patients which may have yet insufficiency to deal properly with the complexities of walking.  Meaningful balance information, in terms of timing and location, can provide this extra signal in any form of stimulation, particularly vibrotactile feedback (VF)~\cite{Islam.etal2022}.  Although the effectiveness of biofeedback on static balance has been studied more extensively in the literature~\cite{DeAngelis.etal2021}, to the best of our knowledge works dealing with dynamic balance problems while walking have been more scarce.

This work presents the development of a device which is grounded on this idea, and aims to help a post-stroke patient with a remaining dynamic balance problem, presenting it as a case study.  The proposed development is implemented as a smart-vest~\cite{Brandebusemeyer.etal2021}, which we will call, \textit{Equilivest}, that address three possible clinical hypothesis of the underlying problem.  We aim to provide motor learning, meaning to generate an assist-as-needed~\cite{Balasubramian.etal2010,Maaref.etal2016} vibrotactile feedback signal which initially promotes voluntary control over movements of the joints~\cite{Islam.etal2022}, but which fades away to encourage automatism~\cite{Srivastava.etal2016,Donato.etal2016}.  The device aims to increase plasticity by producing timing vibrotactile stimulation based on kinematic and dynamics measurements.  




\section{Materials and Methods}

\subsection{Patient Case Study}
\label{sec:case-study}

Patient is a 36 years-old female, without any history of chronic ailment, who suffered an acute brainstem stroke after giving birth.  The stroke was on posterior fossa subarachnoid due to a brain artereovenous malformation (AVM), which could be linked to pregnancy or puerperium~\cite{Porras.etal2017}.  Patient was in induced coma for around 1 month, and after that unable to walk, move, talk or swallow.  Throughout a first period of intensive rehabilitation, the patient managed to recover significantly, including from dysphagia.

After 7 months since event, the patient was discharged from hospital and maintained a daily rehabilitation treatment, focusing on a remaining affection related to dynamic balance problems while walking.  The patient achieved satisfactory index scores in static balance tests and is able to perform static hip-balance and ankle-balance.  She has recovered muscle force in her legs and can perform lower-limb exercises.  Her vision is normal.

However, when the patient tries to walk on open-spaces, or with confronting lights (like walking towards sunlight), with other people moving around, or when concentration fades while walking, she is unable to keep up with the pace of the gait and the risk of fall increases, in a non-lateralized locomotion (symmetric)~\cite{Welniarz.etal2015}. Nowadays, the patient walks freely unaided at home but requires a Canadian cane otherwise and constant supervision.

\subsection{Underlying hypothesis}
\label{sec:hypothesis}

Human balance is composed of a complex interaction of different subsystems, which includes somatosensorial information, vestibular system and visual information as input sources.  These are later processed in different networks of the Central Nervous System (CNS), and finally actuated by motor pathways at many different scales~\cite{Donato.etal2016}.

\begin{table*}[t]
\begin{center}
\begin{tabular}[!t]{|p{0.45\linewidth} |p{0.45\linewidth} |}
\hline
\textbf{Question} & \textbf{Answer} \\
\hline
\hline
Under which situations do you feel you are prone to fall ? & When I am tired, distracted or stressed. \\
\hline
What do you feel ?  & I feel that I loss my balance\\
\hline
What do you feel before you fall ? & Sometimes I do feel something before.  I realize that I am going to fall, but there isn't anything I can do to avoid it. \\
\hline
When do you use the cane ?  & Only when I am out of home.\\
\hline
\end{tabular}
\vspace{2pt}
\caption{Survey and responses provided by the patient.}
\label{tab:patientsurvey}
\end{center}
\end{table*}

\begin{table*}[t]
\begin{center}
\begin{tabular}[!t]{|p{0.45\linewidth} |p{0.45\linewidth} |}
\hline
\textbf{Question} &  \textbf{Answer} \\
\hline
\hline
Describe the situation when the patient loss her balance & Loss of balance is manifested at the beginning, during the gait cycle or with sudden stops. Walking is performed using intense visual compensation.   Instability increase when the patient is distracted, when she stops looking at the floor, or when she looks sideways.  Her gait is not automated, and demands cognitive resources. \\
\hline
Does she presents any of the following muscular dystrophy in lower-limbs / dysmetria / heavy shaking / somatosensorial alterations / proprioceptive alterations ? & No \\
\hline
Does she have any visible reaction before falling?  & She presents dynamic alterations in hip and ankle compensation.\\
\hline
Is the patient gait normal ?  &  No, the patient presents ataxic gait, likely triggered by slow reaction to perform lower-limb balance correction. \\
\hline
Is there any behavioral pattern linked to falling events ? &  Yes, cognitive workload.  Dual tasks situations, when the patient needs to perform something extra while walking. \\
\hline
Describe current treatment. & The patient is currently working on rehabilitation exercises to retrain her Vestibulo-Ocular Reflex (VOR), to improve her static and dynamic balance, and improve her locomotor automatism.  She is improving on a monthly basis, but still she has not reached the level to walk autonomy without any aid. \\
\hline
\end{tabular}
\vspace{2pt}
\caption{Survey and responses provided by patient's rehabilitation caregiver.}
\label{tab:caregiverssurvey}
\end{center}
\end{table*}

We perform a series of surveys and interviews with the patient and their caregivers.  Main results are summarized in 
Tables ~\ref{tab:patientsurvey} and \ref{tab:caregiverssurvey}.  Based on these results and patient's clinical history, we postulate three different potential clinical situations that could use the external biofeedback signal eventually aiding in rehabilitation procedures.  The first (i) is a potential problem in the integration of vestibular information while walking, the second (ii) is bradykinesia where the required processing speed to effectively adjust the lower limb to keep the center of mass inside the base support is not achieved.  Finally, the third (iii) hypothesis is an ataxic gait, where non-automated gait produces an increase in the likelihood of failing.


%

\subsection{Robotic Device Vest}
\label{sec:vest}

\begin{figure}[h!]
\centering
\includegraphics[width=8cm]{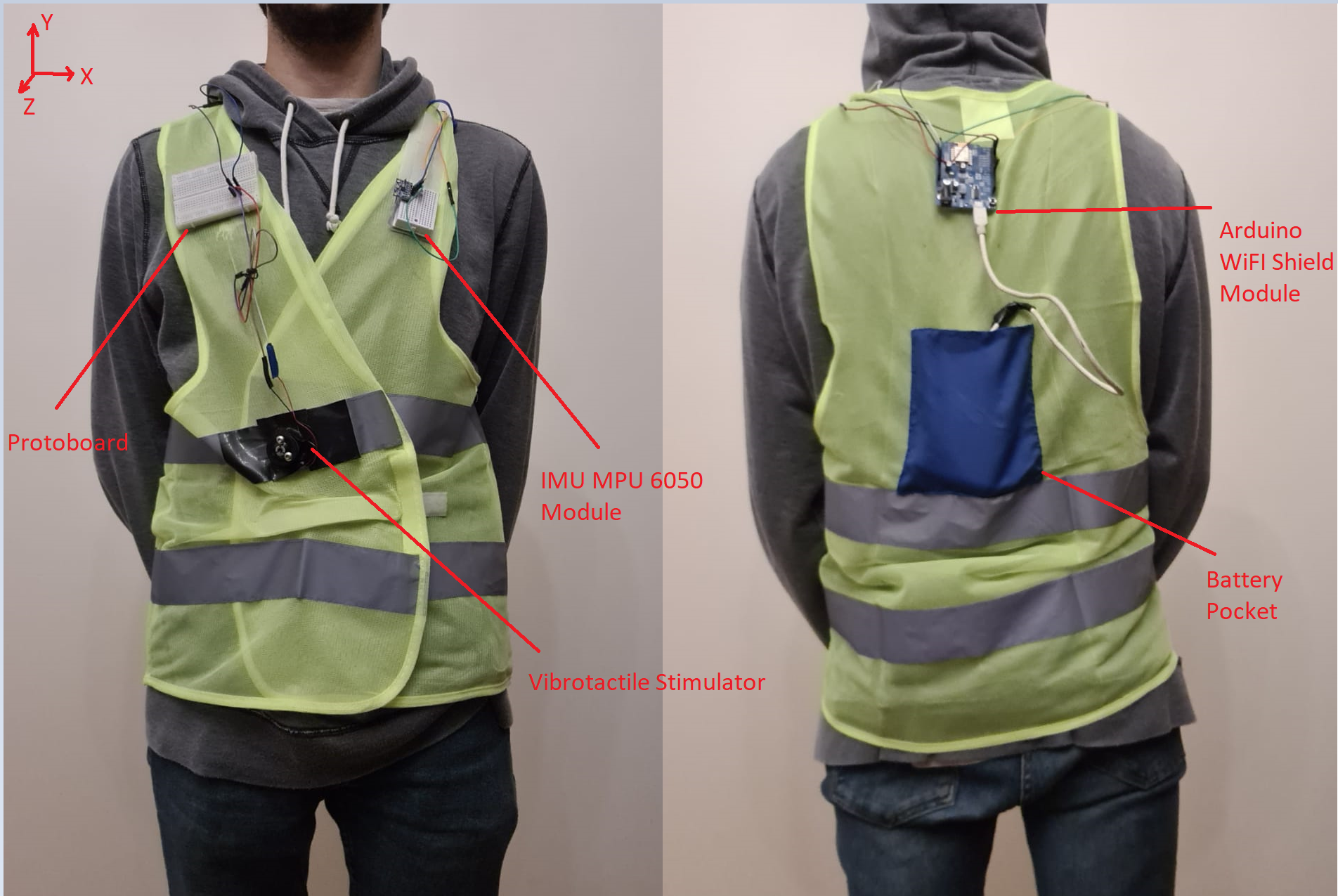}
\caption{Front and back views of the Equilivest prototype.  The vest is based on a safety vest, with additional velcro pockets.  The battery is located on the back, while the vibrotactile motor is located on the belly~\cite{Brandebusemeyer.etal2021}.  The IMU sensor is pressed against the chest by the use of a flexible elastic band. }
\label{fig:smart-vest}
\end{figure}

The vest system prototype is based on Internet of Robot Thing (IoRT) technology~\cite{Simoens.etal2018,Domingo.etal2012}. The controller is an Arduino UNO (Arduino LLC, Italy) board coupled with ESP8266 shield (WeMos, United Kingdom).  The system is powered by a commercial power bank with Li-Po 18650 cells (Ipower, United States). It also contains an Inertial Measurement Unit (IMU) MPU 6050 (OEM ITG/MPU6050) and a FA-12350 DC motor scavenged from old compact discs which provide the vibrotactile feedback.  The IMU provides accelerometer and gyroscope data to get a 3-dimensional angular acceleration vector and a linear acceleration vector. Roll (Z), yaw (Y) and pitch (X) angles, aligned respectively with anatomical planes coronal, horizontal and sagittal are calculated from a set of transforming equations and processed with a complementary filter which relies $0.98$ in gyroscope data and $0.02$ in accelerometer data~\cite{Fetick.2022}. These values are transmitted by telemetry in real-time as UDP packets for off-board register, processing and further analysis. Data recovered by the device is sampled at 100 Hz. The smart-vest prototype can be seen in Figure~\ref{fig:smart-vest}.  It is based on a safety vest, complemented with Velcro pockets. 

\subsection{Experimental Design}
\label{sec:experimental}

Three experiments are designed in order to assess each one of the potential clinical causes, and to derive for each one of them a stimulation strategy.  Participants are recruited voluntarily and the experiment is conducted anonymously in accordance with the declaration of Helsinki published by the World Health Organization. No monetary compensation is handed out and all participants agree and sign a written informed consent. This study is approved by the Departamento de Investigación y Doctorado, Instituto Tecnológico de Buenos Aires (ITBA).  All the participants wear the smart-vest with an elastic band pressed tightly towards the chest, tied in no-restrictive manner, holding the sensor firmly. 


\subsubsection{Vestibular Information Integration}

This experiment aims to study the falling process. Participants are told to lean forward performing ankle forward until they couldn't maintain balance anymore and let themselves fall into a mattress. This experiment is used to determine a threshold for the data that can help to identify the breakpoint conditions based on the IMU information where each subject falls.

In order to test it, 5 healthy participants are recruited to perform 10 runs of falling situations  Participants wearing the vest lean forward by adjusting their ankle joint, without performing a one-step forward movement,  at different angles progressively until they can no longer cope with the unbalance situation and fall to the mattress.

\subsubsection{Ataxic Gait}

This experiment has the purpose of analyzing and studying gait's behavior. In order to accomplish this,  5 participants are instructed to execute 10 sessions of walking in a straight line across ten meters, performing Ten-Meter Walking (10MWT) assesment~\cite{Olmos.etal2008}.  We collected and analyzed pitch, roll and yaw values as well as angular acceleration changes registered by the IMU gyroscope. In this study our goal is to be able to process and identify gait abnormalities as well as all steps occurring and any fall that might develop.

\subsubsection{Bradykinesia}

This experiment involves the coupling of the other two.  Five participants are instructed to perform 10 runs of a ten-meter walking procedure, followed by a falling into a mattress.  The purpose of this experiment is obtain a multichannel time series of the whole sequence, including the walk and the falling action.



\section*{Results and Discussion}

\begin{figure}[h!]
\centering
\includegraphics[width=8cm]{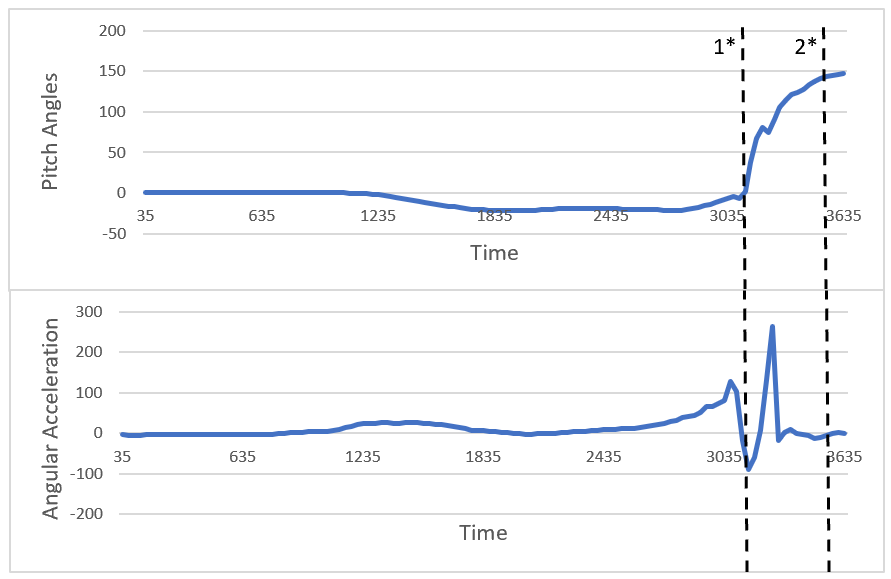}
\caption{Pitch value in degrees and angular acceleration around X axis showing the progressive increase as participants leaned forward crossing the instability condition (1).  Time is in milliseconds. }
\label{fig:vestibular}
\end{figure}

\begin{figure}[h!]
\centering
\includegraphics[width=8cm]{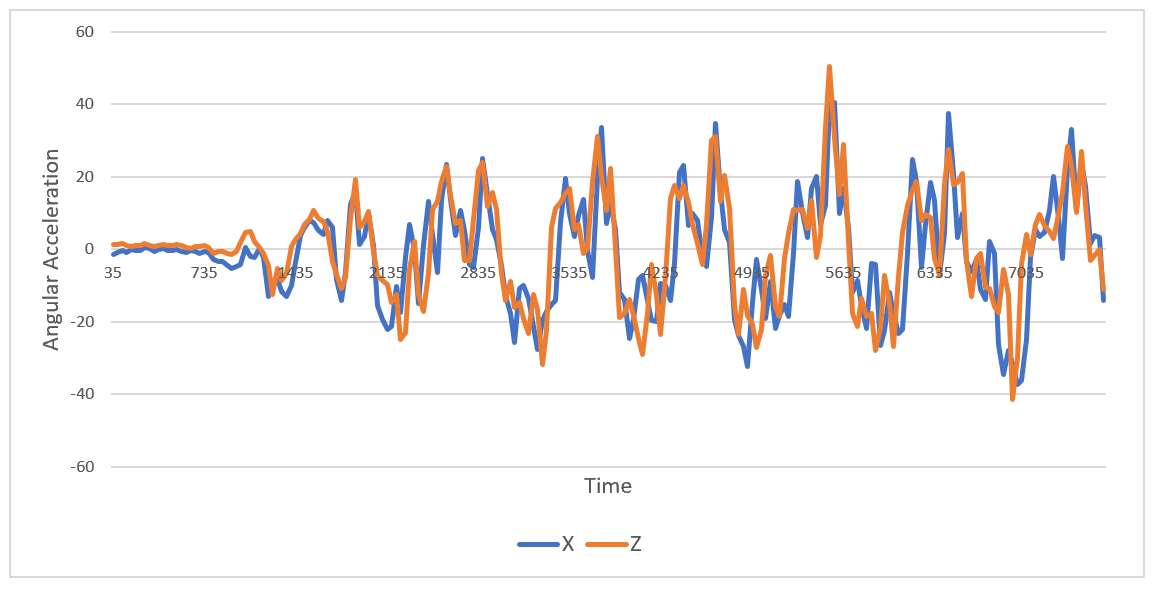}
\caption{Angular acceleration for the X and Z axis showing 7 steps of the 10MWT assessment.  Time is in milliseconds. }
\label{fig:gait}
\end{figure}

Preliminary results show that for the first (1) experiment, on Figure \ref{fig:vestibular}, the pitch angle can be used to determine a breakpoint where the fall is inevitable (black vertical line, marked as 1*).   On the other hand, results from experiment number (2) in Figure \ref{fig:gait} show that the gait cycle is clearly visible from raw acceleration data.

%
%
%

The information provided from all the experiments allow us to implement three different VF strategies:  The first is an \textbf{Artificial Vestibular Feedback}, which provides information as a feedback for vestibular information integration process.  This is implemented as a frequency-increasing VF signal that increases as long as the distance from the breaking condition position where the fall is inevitable, decreases.   The purpose is to provide the patient with a continuous sensation that modulates the risk of falling.

The second strategy, independent of the first one, is the implementation of \textbf{Gait Pacemaker}.  The obtained data from the analysis of the ataxic gait is used to implement a podometer.  It has been showed that gait synchronization with music increase gait performance~\cite{Roerdink.etal2007} and synchronized gait trainer achieved positive results on patient~\cite{Blicher.etal2009}.  Hence, the aim of this stimulation approach is to keep the patient as close as possible to a normal and safe gait.

The final stimulation scheme deals with the Bradykinesia condition, and aims to have a \textbf{Risk-Predictor}~\cite{Rahman.etal2022,Ali.etal2022}.  The IMU data  represents a multichannel time series.  Hence, it can be used to predict a risk of falling, ahead of its occurrence, by building a machine-learning predictor based on the analysis of the time-series data.  The rationale is to counter the slowness of the response by predicting the falling situation ahead of its occurrence and providing the VF stimulation to force a change in the risky gait pattern.



\section*{Conclusion}
Overall, we aim for \textit{Equilivest} to serve as a testbed that can be iteratively extended to obtain experimental data and to implement biofeedback strategies~\cite{Bowman2021}.



%
%
%

\bibliographystyle{IEEEtran}
\bibliography{iros}

\begin{thebibliography}{10}
\providecommand{\url}[1]{#1}
\csname url@samestyle\endcsname
\providecommand{\newblock}{\relax}
\providecommand{\bibinfo}[2]{#2}
\providecommand{\BIBentrySTDinterwordspacing}{\spaceskip=0pt\relax}
\providecommand{\BIBentryALTinterwordstretchfactor}{4}
\providecommand{\BIBentryALTinterwordspacing}{\spaceskip=\fontdimen2\font plus
\BIBentryALTinterwordstretchfactor\fontdimen3\font minus
  \fontdimen4\font\relax}
\providecommand{\BIBforeignlanguage}[2]{{%
\expandafter\ifx\csname l@#1\endcsname\relax
\typeout{** WARNING: IEEEtran.bst: No hyphenation pattern has been}%
\typeout{** loaded for the language `#1'. Using the pattern for}%
\typeout{** the default language instead.}%
\else
\language=\csname l@#1\endcsname
\fi
#2}}
\providecommand{\BIBdecl}{\relax}
\BIBdecl

\bibitem{Caplan.etal2023}
\BIBentryALTinterwordspacing
L.~R. Caplan, R.~P. Simon, and S.~Hassani, ``Chapter 27 - cerebrovascular
  disease—stroke,'' in \emph{Neurobiology of Brain Disorders (Second
  Edition)}, second edition~ed., M.~J. Zigmond, C.~A. Wiley, and M.-F.
  Chesselet, Eds.\hskip 1em plus 0.5em minus 0.4em\relax Academic Press, 2023,
  pp. 457--476. [Online]. Available:
  \url{https://www.sciencedirect.com/science/article/pii/B9780323856546000447}
\BIBentrySTDinterwordspacing

\bibitem{Bowman2021}
T.~Bowman, E.~Gervasoni, C.~Arienti, S.~G. Lazzerini, S.~Negrini, S.~Crea,
  D.~Cattaneo, and M.~C. Carrozza, ``{Wearable devices for biofeedback
  rehabilitation: A systematic review and meta-analysis to design application
  rules and estimate the effectiveness on balance and gait outcomes in
  neurological diseases},'' \emph{Sensors}, vol.~21, no.~10, 2021.

\bibitem{Afrasiabifar.etal2020}
\BIBentryALTinterwordspacing
A.~Afrasiabifar, Z.~Mehri, and H.~R.~G. Shirazi, ``Orem’s self-care model
  with multiple sclerosis patients’ balance and motor function,''
  \emph{Nursing Science Quarterly}, vol.~33, no.~1, pp. 46--54, 2020, pMID:
  31795883. [Online]. Available: \url{https://doi.org/10.1177/0894318419881792}
\BIBentrySTDinterwordspacing

\bibitem{Donato.etal2016}
\BIBentryALTinterwordspacing
S.~M. Donato, K.~H. Pulaski, and G.~Gillen, ``Chapter 19 - overview of balance
  impairments: Functional implications,'' in \emph{Stroke Rehabilitation
  (Fourth Edition)}, fourth edition~ed., G.~Gillen, Ed.\hskip 1em plus 0.5em
  minus 0.4em\relax Mosby, 2016, pp. 394--415. [Online]. Available:
  \url{https://www.sciencedirect.com/science/article/pii/B9780323172813000198}
\BIBentrySTDinterwordspacing

\bibitem{DeAngelis.etal2021}
S.~{De Angelis}, A.~A. Princi, F.~{Dal Farra}, G.~Morone, C.~Caltagirone, and
  M.~Tramontano, ``{Vibrotactile-based rehabilitation on balance and gait in
  patients with neurological diseases: A systematic review and metanalysis},''
  \emph{Brain Sciences}, vol.~11, no.~4, 2021.

\bibitem{Albert.etal2012}
\BIBentryALTinterwordspacing
S.~J. Albert and J.~Kesselring, ``{Neurorehabilitation of stroke},''
  \emph{Journal of Neurology}, vol. 259, no.~5, pp. 817--832, 2012. [Online].
  Available: \url{https://doi.org/10.1007/s00415-011-6247-y}
\BIBentrySTDinterwordspacing

\bibitem{Peake.etal2018}
J.~M. Peake, G.~Kerr, and J.~P. Sullivan, ``A critical review of consumer
  wearables, mobile applications, and equipment for providing biofeedback,
  monitoring stress, and sleep in physically active populations,''
  \emph{Frontiers in physiology}, vol.~9, p. 743, 2018.

\bibitem{Zhao.etal2022}
\BIBentryALTinterwordspacing
A.~D. Zhao, B.~T. Zhang, C.~H. Liu, D.~J. Yang, and E.~H. Yokoi, ``Gait
  rehabilitation training robot: A motion-intention recognition approach with
  safety and convenience,'' \emph{Robotics and Autonomous Systems}, p. 104260,
  2022. [Online]. Available:
  \url{https://www.sciencedirect.com/science/article/pii/S092188902200152X}
\BIBentrySTDinterwordspacing

\bibitem{Peshkin.etal2005}
M.~Peshkin, D.~A. Brown, J.~J. Santos-Munn{\'e}, A.~Makhlin, E.~Lewis, J.~E.
  Colgate, J.~Patton, and D.~Schwandt, ``Kineassist: A robotic overground gait
  and balance training device,'' in \emph{9th International Conference on
  Rehabilitation Robotics, 2005. ICORR 2005.}\hskip 1em plus 0.5em minus
  0.4em\relax IEEE, 2005, pp. 241--246.

\bibitem{Tong.etal2006}
R.~K. Tong, M.~F. Ng, and L.~S. Li, ``Effectiveness of gait training using an
  electromechanical gait trainer, with and without functional electric
  stimulation, in subacute stroke: a randomized controlled trial,''
  \emph{Archives of physical medicine and rehabilitation}, vol.~87, no.~10, pp.
  1298--1304, 2006.

\bibitem{Brandebusemeyer.etal2021}
C.~Brandebusemeyer, A.~R. Luther, S.~U. K{\"{o}}nig, P.~K{\"{o}}nig, and S.~M.
  K{\"{a}}rcher, ``{Impact of a vibrotactile belt on emotionally challenging
  everyday situations of the blind},'' \emph{Sensors}, vol.~21, no.~21, 2021.

\bibitem{Islam.etal2022}
\BIBentryALTinterwordspacing
M.~S. Islam and S.~Lim, ``Vibrotactile feedback in virtual motor learning: A
  systematic review,'' \emph{Applied Ergonomics}, vol. 101, p. 103694, 2022.
  [Online]. Available:
  \url{https://www.sciencedirect.com/science/article/pii/S0003687022000175}
\BIBentrySTDinterwordspacing

\bibitem{Balasubramian.etal2010}
S.~Balasubramanian, H.~Zhang, S.~Buchanan, H.~Austin, R.~Herman, and J.~He,
  ``Cooperative and active assistance based interactive therapy,'' in
  \emph{IEEE/ICME International Conference on Complex Medical
  Engineering}.\hskip 1em plus 0.5em minus 0.4em\relax IEEE, 2010, pp.
  311--315.

\bibitem{Maaref.etal2016}
M.~Maaref, A.~Rezazadeh, K.~Shamaei, R.~Ocampo, and T.~Mahdi, ``A bicycle
  cranking model for assist-as-needed robotic rehabilitation therapy using
  learning from demonstration,'' \emph{IEEE Robotics and Automation Letters},
  vol.~1, no.~2, pp. 653--660, 2016.

\bibitem{Srivastava.etal2016}
\BIBentryALTinterwordspacing
S.~Srivastava and P.~C. Kao, ``{Robotic Assist-As-Needed as an Alternative to
  Therapist-Assisted Gait Rehabilitation},'' \emph{International Journal of
  Physical Medicine \& Rehabilitation}, vol.~4, no.~5, 2016. [Online].
  Available: \url{https://www.ncbi.nlm.nih.gov/pmc/articles/PMC5450822/}
\BIBentrySTDinterwordspacing

\bibitem{Porras.etal2017}
J.~L. Porras, W.~Yang, E.~Philadelphia, J.~Law, T.~Garzon-Muvdi, J.~M. Caplan,
  G.~P. Colby, A.~L. Coon, R.~J. Tamargo, and J.~Huang, ``Hemorrhage risk of
  brain arteriovenous malformations during pregnancy and puerperium in a north
  american cohort,'' \emph{Stroke}, vol.~48, no.~6, pp. 1507--1513, 2017.

\bibitem{Welniarz.etal2015}
Q.~Welniarz, I.~Dusart, C.~Gallea, and E.~Roze, ``One hand clapping:
  lateralization of motor control,'' \emph{Frontiers in neuroanatomy}, vol.~9,
  p.~75, 2015.

\bibitem{Simoens.etal2018}
\BIBentryALTinterwordspacing
P.~Simoens, M.~Dragone, and A.~Saffiotti, ``The internet of robotic things: A
  review of concept, added value and applications,'' \emph{International
  Journal of Advanced Robotic Systems}, vol.~15, p.~10, 1 2018. [Online].
  Available: \url{http://journals.sagepub.com/doi/10.1177/1729881418759424}
\BIBentrySTDinterwordspacing

\bibitem{Domingo.etal2012}
\BIBentryALTinterwordspacing
M.~C. Domingo, ``An overview of the internet of things for people with
  disabilities,'' \emph{Journal of Network and Computer Applications}, vol.~35,
  pp. 584--596, 3 2012. [Online]. Available:
  \url{https://www.sciencedirect.com/science/article/pii/S1084804511002025}
\BIBentrySTDinterwordspacing

\bibitem{Fetick.2022}
R.~J. Fetick, \emph{MPU6050 Library}, 2022 (accessed September 16, 2022),
  \url{https://github.com/rfetick/MPU6050\_light}.

\bibitem{Olmos.etal2008}
\BIBentryALTinterwordspacing
L.~E. Olmos, O.~Freixes, M.~A. Gatti, D.~A. Cozzo, S.~A. Fernandez, C.~J. Vila,
  P.~E. Agrati, and I.~F. Rubel, ``{Comparison of gait performance on different
  environmental settings for patients with chronic spinal cord injury},''
  \emph{Spinal Cord}, vol.~46, no.~5, pp. 331--334, 2008. [Online]. Available:
  \url{https://doi.org/10.1038/sj.sc.3102132}
\BIBentrySTDinterwordspacing

\bibitem{Roerdink.etal2007}
M.~Roerdink, C.~J. Lamoth, G.~Kwakkel, P.~C. Van~Wieringen, and P.~J. Beek,
  ``Gait coordination after stroke: benefits of acoustically paced treadmill
  walking,'' \emph{Physical therapy}, vol.~87, no.~8, pp. 1009--1022, 2007.

\bibitem{Blicher.etal2009}
J.~U. Blicher and J.~F. Nielsen, ``Cortical and spinal excitability changes
  after robotic gait training in healthy participants,''
  \emph{Neurorehabilitation and neural repair}, vol.~23, no.~2, pp. 143--149,
  2009.

\bibitem{Rahman.etal2022}
\BIBentryALTinterwordspacing
W.~Rahman, M.~Hasan, M.~S. Islam, T.~Olubajo, J.~Thaker, A.~Abdelkader,
  P.~Yang, T.~Ashizawa, and E.~Hoque, ``Auto-gait: Automatic ataxia risk
  assessment with computer vision on gait task videos,'' 2022. [Online].
  Available: \url{https://arxiv.org/abs/2203.08215}
\BIBentrySTDinterwordspacing

\bibitem{Ali.etal2022}
\BIBentryALTinterwordspacing
F.~Ali and E.~Benarroch, ``What is the brainstem control of locomotion?''
  \emph{Neurology}, vol.~98, no.~11, pp. 446--451, 2022. [Online]. Available:
  \url{https://n.neurology.org/content/98/11/446}
\BIBentrySTDinterwordspacing

\end{thebibliography}

%

\end{document}